# Multi-Layer Context Camouflaging: A Semantic Superposition and Contextual Lamination Framework for Malpractice-Resilient Online Assessment

Lovi Raj Gupta[1],*, Kamalpreet Kaur[1], Sri Ram[1], Prajithaa[1]
[1]*Lovely Professional University, Punjab, India*
**Corresponding author: loviraj@gmail.com*

***Abstract***

This paper extends the Multi-dimensional Spatio-Temporal Context Camouflaging Model (MSCCM) introduced in prior work on the MARS assessment-resilience suite, developing a substantially more detailed mathematical treatment of its Context Camouflaging Operator. We formalize Context Camouflaging as a semantic superposition process, termed the Multi-Layer Context Camouflaging Theory (MCCT), in which an authentic semantic stream and a generated camouflage stream coexist within a single rendered surface without ordinary concatenation.

The revised formulation separates the rendered surface from the sequence an adversary actually recovers, and defines an extraction-channel operator that makes explicit which attributes (glyph, colour, position, time) survive each capture route. Against that model we develop six coupled constructs: a Context Inversion Operator with locality and vocabulary-closure conditions; a Contextual Lamination Operator driven by a keyed insertion set; a Separation Channel that generalizes colour to any modality carrying a perceptual discriminability margin; a Human Readability Functional conditioned on that margin; a Computational Ambiguity Functional redefined as the conditional entropy of the authentic stream given the adversary view, for which we derive the closed form $A_c = \log_2 C(n+m, m)$; and a Context Camouflage Tensor coupling the scheme to MARS temporal rendering.

We also correct the direction of the preservation constraint. Recoverability for the legitimate viewer is an exact filtering identity rather than a similarity bound, so the operative constraints become a fidelity identity, an obfuscation ceiling on Sim(Q, Ω), and a plausibility ceiling on the detectability of camouflage tokens under a language-model prior. Eight theorems follow, including a closed-form ambiguity result, a corrected optimal-density result with a binding-constraint corollary, a semantic-filter degradation bound, a multi-observation leakage theorem showing that frame-granular re-lamination is unsafe against a multi-capture adversary, and a coupon-collector bound on capture complexity under temporal multiplexing. We close with an algorithm, its complexity, and a pre-registered evaluation protocol. No empirical results are claimed.

## I. INTRODUCTION

Online assessment platforms remain vulnerable to a class of malpractice in which the rendered content of an assessment item, not merely the candidate's behaviour around it, is extracted by screenshot, copy-paste, or automated scraping and relayed to an unauthorized solver, whether human or algorithmic. Prior work on the Multi-modal Assessment Resilience Suite (MARS) introduced Context Camouflaging as one operator within a broader six-dimensional dynamical model of assessment integrity, defined at the level of a single transformation Φ mapping an original question Q to a semantically equivalent rendering Q′ [1]. That formulation established the existence of the operator and a coarse semantic-similarity constraint, but did not develop the internal structure of Φ in mathematical depth.

This paper develops that internal structure. We show that Context Camouflaging is best understood not as a single content-rewriting step but as a semantic superposition: the authentic content and a separately generated camouflage stream are rendered together on a shared surface, laminated at keyed positions, and separated perceptually so that a human and an automated extraction pipeline read different documents from the same pixels.

Three things changed in the course of that development, and it is worth stating them plainly at the outset rather than burying them in the analysis. First, a scheme of this kind cannot be evaluated without naming the adversary. Colour survives a DOM scrape and dies in a clipboard copy, and a scheme whose whole security rests on colour therefore has entirely different properties against the two. Section III-B introduces an explicit extraction-channel model and a five-class adversary taxonomy, and every later claim is stated relative to a named class.

Second, the preservation constraint in [1] pointed the wrong way. Requiring Sim(Q, Ω) ≥ θ asks the laminated field to resemble the authentic item, which is precisely what makes it useful to a solver holding the extracted text. The legitimate viewer does not need similarity at all, because filtering by the separation channel returns Q exactly. Section IX replaces the single similarity floor with a fidelity identity, an obfuscation ceiling, and a plausibility ceiling, which is the constraint set the design actually has to satisfy.

Third, dynamic lamination as originally described leaks. If the authentic tokens stay fixed while the camouflage layer is resampled every frame, then across a handful of captures the authentic tokens are simply the ones that never change. Theorem 7 quantifies the decay and Section X gives the condition under which temporal variation is safe. This result runs against the intuition that more randomization means more security, and it is the finding we would most want a reviewer to check.

Contributions. (i) An explicit extraction-channel model and adversary taxonomy for rendered-content attacks on assessment items. (ii) A closed-form expression for computational ambiguity as conditional entropy, $A_c = \log_2 C(n+m,\ m)$, replacing the unigram-entropy difference used previously. (iii) A corrected three-part constraint set with a Lagrangian treatment of camouflage density. (iv) A generalized separation channel with a perceptual discriminability margin expressed in CIEDE2000 units, together with a colour-vision-deficiency condition. (v) Eight theorems, of which four are new, including the multi-observation leakage result and a capture-complexity bound under temporal multiplexing. (vi) A stated algorithm with complexity analysis and a pre-registered evaluation protocol.

Sections II and III cover related work, the MARS recap, and the threat model. Sections IV to XI develop the constructs. Section XII proves the eight theorems, Section XIII gives the algorithm, Section XIV the evaluation protocol, and Sections XV and XVI discuss limitations and conclude.

## II. Related Work

This paper is a direct mathematical extension of the MARS assessment-resilience framework, in which assessment integrity is modelled as a coupled dynamical system rather than a set of independently engineered safeguards, and the Context Camouflaging Operator is introduced as one of its six coupled constructs [1]. The present work develops that operator's internal mathematics without altering the surrounding system model. Dawson documents the limitations of treating browser lockdown, webcam surveillance, and behavioural analytics as independent, uncoordinated safeguards, motivating systems that address the rendered-content channel directly rather than only the candidate's behaviour around it [2].

The mechanism exploited here, a deliberate gap between what a human sees and what a machine parses, has a direct precedent in security research that we did not cite in the earlier version and should have. Boucher et al. showed that Unicode homoglyphs, invisible characters, and bidirectional reordering let an attacker construct text whose visual rendering and logical encoding disagree, and used that gap to attack NLP pipelines in a black-box setting [7]. Boucher and Anderson extended the same idea to source code, where the compiler and the reviewer read different programs from one file [8]. MCCT inverts the polarity of that attack: the same rendering-versus-encoding gap is used defensively, and the party disadvantaged by the gap is the extraction pipeline rather than the human. Reading the two literatures together also sets a useful expectation. Those attacks were eventually mitigated by input sanitization at the parser, and the analogous countermeasure here is an adversary that normalizes the rendered surface before parsing, which is exactly the A4 and A5 classes of Section III-B.

The claim that a human can filter a colour-separated stream at negligible cost rests on visual-search results rather than on assertion. Treisman and Gelade established that a target differing from distractors in a single primitive feature such as hue is detected in time roughly independent of the number of distractors, and Wolfe's Guided Search work refines the conditions under which that independence holds and where it breaks down [9], [10]. Both point to the same practical requirement: the separation must be a single preattentive feature with an adequate discriminability margin, otherwise search becomes serial and readability degrades with camouflage density. We express that margin in CIEDE2000 units [11], with the commonly cited average just-noticeable difference of roughly 2.3 CIELAB units as the lower reference point [12]. Colour-vision deficiency is handled by requiring the margin to survive dichromatic simulation [13], or by moving the separation to a non-chromatic modality.

The Context Inversion Operator generalizes a simple negation-token substitution into an operator that draws on lexical-semantic resources; WordNet's synonym and antonym relations provide one practical basis for generating context-dependent, polarity-reversing substitutions without relying on a fixed vocabulary list [3]. Computing the semantic similarity that appears in the obfuscation constraint can be operationalized with transformer-based sentence encoders: BERT provides contextual token representations from which sentence-level meaning can be derived [4], and Sentence-BERT adapts this into an efficient, directly comparable sentence-embedding form [5]. Shannon's information theory supplies the entropy formalism used throughout to quantify computational ambiguity and to bound the behaviour of dynamic lamination [6].

What remains unaddressed in the prior literature, and what this paper supplies, is a treatment of the perceptual gap as a designed defensive primitive with a stated adversary model, a closed-form ambiguity measure, and an account of how the guarantee degrades against a semantically informed attacker.

## III. MARS System Model and Threat Model

### *A. Recap of the MARS state*

MARS represents an assessment session as the coupled state tuple

$$M = \langle\, S(t), T(t), C(t), B(t), R(t), I(t) \,\rangle \qquad (1)$$

comprising spatial, temporal, contextual, behavioural, rendering, and integrity components [1]. The present paper is concerned with the contextual domain C(t) and its associated Context Camouflaging Operator Φ, previously defined at the coarse level $Q' = \Phi(Q,S,T,C)$ subject to $\mathrm{sim}(Q,Q') \geq \theta$. Everything that follows refines this single operator into the Multi-Layer Context Camouflaging Theory (MCCT) of Sections IV to XI, culminating in a unified rendering equation (Section XI) that reconnects MCCT to the rendering dynamics established in [1].

One notational change is needed before proceeding. The symbol C was used in the earlier version for the contextual state C(t), the colour assignment C(w), and the binomial coefficient C(n+m, m). We retain C(t) and $C(\cdot,\cdot)$ and rename the colour assignment to $\chi(\cdot)$.

### *B. Extraction channels and adversary classes*

A rendered assessment item is not a token sequence. It is a set of glyphs carrying position, colour, and a time index, and the question of what an attacker obtains is the question of which of those attributes survive the capture route. Writing the rendered surface as

$$\Sigma(t) = (\sigma_1, \ldots, \sigma_N), \quad \sigma_i = (\Omega_i, x_i, y_i, c_i, \tau_i), \quad N = n + m \tag{2}$$

an extraction channel is a projection

$$X_A : \Sigma(t_{1:k}) \mapsto V_A, \quad V_A = X_A(\Sigma(t_1), \ldots, \Sigma(t_k)) \tag{3}$$

where $V_A$ is the adversary view and k the number of captures. Table I records which attributes each channel preserves. The classes are ordered by increasing capability, and every security claim later in the paper is stated against a named class rather than against an unspecified attacker.

Two entries deserve emphasis because they bound what the scheme can honestly claim. Class A3, a scripted reader of the document object model, sees the styling that defines the separation channel; colour separation alone therefore provides no protection against A3 unless the styling is rendered non-recoverable, for example by drawing to a canvas or by randomizing per-token class names so that authentic and camouflage tokens are not distinguishable by selector. Class A5, a vision-language model given a screenshot, also sees colour, and additionally supplies the semantic prior that makes the combinatorial bound of Theorem 5 collapse. Colour separation is a defence against A1 and A2. Against A3 to A5 the defence has to come from elsewhere in the MARS stack, principally from the temporal visibility term of Section XI, and Theorem 8 states what that buys.

***Table I***
**EXTRACTION CHANNELS AND ATTRIBUTE SURVIVAL**

| Class | Channel | Glyph | Colour | Position | Semantics |
|---|---|---|---|---|---|
| A1 | Clipboard copy to plain text | yes | no | order only | none |
| A2 | Screenshot then OCR | yes | no | yes | none |
| A3 | Scripted DOM reader | yes | yes | yes | none |
| A4 | A1 or A2 plus a language-model filter | yes | no | yes | yes |
| A5 | Screenshot to a vision-language model | yes | yes | yes | yes |

*Attribute survival by capture route. "Semantics" denotes whether the adversary can score candidate reconstructions for linguistic plausibility. Colour separation is effective against A1 and A2 only.*

Following Kerckhoffs's principle we assume throughout that the adversary knows the algorithm, the inversion operator, the density η, and the distribution from which insertion sets are drawn. Only the per-session key is secret. Claims that depend on the adversary not knowing the construction are not made.

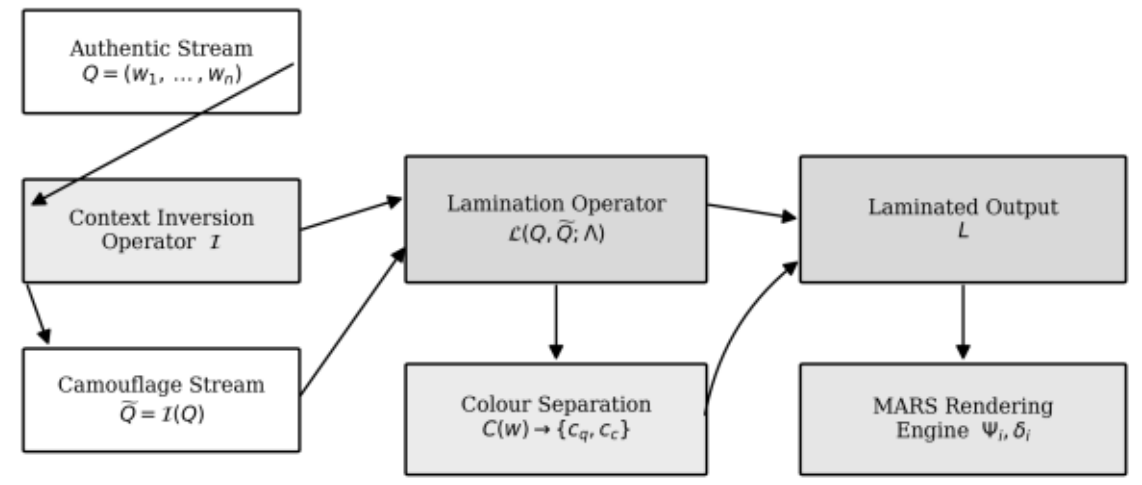

*Fig. 1. Architecture of the Multi-Layer Context Camouflaging Theory (MCCT): the authentic stream and its inversion-generated camouflage stream are laminated, separated by channel, and passed to the MARS rendering engine.*

## IV. Semantic Superposition and the Rendered Surface

Rather than presenting an assessment item as a single token sequence, MCCT renders a superposed linguistic field in which two token streams occupy the same rendered surface, only one of which contributes to the intended meaning. The authentic semantic stream is

$$Q = (w_1, w_2, \ldots, w_n), \quad w_i \in V \tag{4}$$

where V denotes the assessment-item vocabulary. A separately generated camouflage stream

$$\tilde{Q} = (\tilde{w}_1, \tilde{w}_2, \ldots, \tilde{w}_m) \tag{5}$$

is produced by the Context Inversion Operator of Section V. The two streams are combined, not by concatenation but by the lamination process of Section VI, into a superposed field

$$\Omega = \mathcal{L}(Q, \tilde{Q}; \Lambda) \tag{6}$$

The distinction that the earlier version left implicit is between Ω and Σ. The laminated field Ω is a token sequence; the rendered surface Σ of Equation (2) is Ω decorated with position, colour, and visibility. A legitimate candidate observes Σ. An adversary observes $X_A(\Sigma)$, which for classes A1 and A2 is Ω alone. The defining property of the construction can now be stated exactly rather than informally: there exists a filter $\mathcal{F}$ such that $\mathcal{F}(\Sigma; \Gamma) = Q$ with no error at all, while Q is not determined by Ω. Sections V to VII build the three mechanisms that produce this asymmetry and Sections VIII and IX quantify it.

## V. The Context Inversion Operator

The camouflage stream is generated by a Context Inversion Operator $\mathcal{I}$ acting on the authentic stream,

$$\tilde{Q} = \mathcal{I}(Q) \tag{7}$$

Following the generalization principle adopted for this framework, $\mathcal{I}$ produces a context-preserving, computationally ambiguous overlay rather than a fixed vocabulary substitution. One embodiment realizes $\mathcal{I}$ using polarity-reversing tokens ("NOT", "EXCEPT", "UNEQUAL", "UNLESS"); another may draw on qualifiers, antonyms, distractor clauses, or domain-specific semantic modifiers, for instance using the lexical antonym and hypernym relations catalogued in WordNet [3].

The earlier version described what $\mathscr{J}$ should do in prose. Stating it as three conditions makes the later theorems checkable. Let $s(j) \subseteq \{1, \ldots, n\}$ be the source span from which the j-th camouflage token is derived. Then $\mathscr{J}$ is admissible when

$$\text{(C1) locality:}\ \ \tilde{w}_j = g(\, w_i : i \in s(j)\,),\ \ |s(j)| \le \ell \qquad (8)$$

$$\text{(C2) vocabulary closure:}\ \ \min_{v \in V} \mathrm{dist}(\tilde{w}_j, v) \le \rho_V \qquad (9)$$

$$\text{(C3) polarity reversal:}\ \ \langle e(\tilde{w}_j), e(w_{s(j)}) \rangle \le -\kappa_p \qquad (10)$$

where $e(\cdot)$ is a sentence-encoder embedding, dist a metric on that embedding space, $\ell$ a span-width bound, and $\rho_V$ and $\kappa_p$ tunable margins. Condition C1 keeps the operator local, so that $\tilde{Q}$ is anchored to Q span by span rather than to the global meaning of Q. Condition C2 keeps camouflage tokens close to the item vocabulary, which is what prevents an adversary from separating the streams by a vocabulary test alone. Condition C3 is what stops $\tilde{Q}$ from being a usable paraphrase of Q, and therefore what stops the laminated field from leaking the answer to a solver that reads it whole.

C2 and C3 pull against each other, and the tension is real rather than an artefact of the formalism. Pushing $\kappa_p$ up drives the camouflage tokens away from the item vocabulary and makes them easier to spot; pushing $\rho_V$ down brings them back into the vocabulary and weakens the polarity reversal. The plausibility constraint of Section IX is where that trade-off is made explicit.

## VI. The Contextual Lamination Operator

The defining departure from ordinary text insertion is that Q and $\tilde{Q}$ are combined by lamination rather than concatenation. The operator

$$\Omega = \mathscr{L}(Q, \tilde{Q}; \Lambda) \qquad (11)$$

is governed by a control law

$$\Lambda = (\pi, \kappa, \chi) \qquad (12)$$

where $\pi$ is the insertion position map, $\kappa$ the insertion density, and $\chi$ the separation-channel assignment of Section VII. The map $\pi$ determines an insertion set

$$\Pi \subseteq \{1, \ldots, n + m\},\ \ |\Pi| = m,\ \ \kappa = m / (n + m) \qquad (13)$$

and the laminated field is defined position by position as

$$\Omega_i = w_{r(i)} \text{ if } i \notin \Pi;\ \ \tilde{w}_{j(i)} \text{ if } i \in \Pi \qquad (14)$$

where r(i) and j(i) index the authentic and camouflage tokens occupying laminated position i, both order-preserving. Because $\Pi$ is chosen independently of concatenation order, $\Omega$ cannot be separated into its constituent streams by position alone; recovering Q from $\Omega$ requires knowledge of $\Pi$, formalized as Theorem 5.

Where $\Pi$ comes from was left open previously, and it matters, because a $\Pi$ that is merely "random" gives no reproducibility across the render path and no rotation policy across sessions. We derive it from a keyed pseudorandom function,

$$\Pi = \Psi_{PRF}(K_{sess}, id, 0),\ \ \ \Gamma = (c_q, c_c) = \Psi_{PRF}(K_{sess}, id, 1) \qquad (15)$$

with $K_{sess}$ a per-session key held by the rendering service with 0 and 1 acting as domain separators. The pair $(\Pi, \Gamma)$ is the session rendering key. Sampling $\Pi$ uniformly among the m-subsets of $\{1, \ldots, n+m\}$ is what makes the uniform prior of Theorem 3 the correct one; any biased sampler reduces the adversary's uncertainty below the closed form and should be treated as a defect rather than a tuning choice.

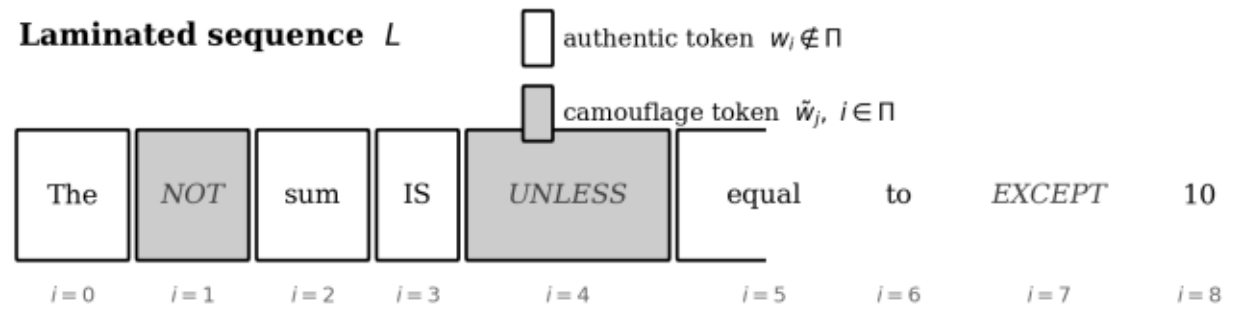

*Fig. 2. Contextual lamination of an authentic stream (white) with inversion-generated camouflage tokens (shaded) at keyed insertion set Π.*

## VII. The Separation Channel and Perceptual Discriminability

Humans exploit colour as an immediate preattentive grouping cue; extraction pipelines in classes A1 and A2 tokenize rendered or copied text after formatting metadata has been discarded by the copy, OCR, or scrape operation. MCCT exploits this asymmetry through a separation function

$$\chi(i) = c_q \text{ if } i \notin \Pi;\ \ c_c \text{ if } i \in \Pi \qquad (16)$$

which assigns the authentic-stream value $c_q$ to positions drawn from Q and the camouflage value $c_c$ to positions drawn from $\tilde{Q}$. The pair $\Gamma = (c_q, c_c)$, together with $\Pi$, constitutes the session rendering key required to invert the lamination, and $\Gamma$ may be rotated across sessions in the same manner as other MARS session parameters.

Treating $\chi$ as specifically chromatic was an unnecessary restriction. What the construction requires of the separation channel is only that it be a single preattentive feature carrying a discriminability margin above threshold, so we state the requirement rather than the mechanism. For a chromatic channel the condition is

$$\Delta E_{00}(\, c_q, c_c\,) \ge \Delta E^*,\ \ \Delta E^* \gg 2.3 \qquad (17)$$

with $\Delta E_{00}$ the CIEDE2000 colour difference [11] and 2.3 CIELAB units the commonly cited average just-noticeable difference [12]. A margin near the JND is the wrong operating point, because the readability result of Theorem 2 depends on the difference being preattentive rather than merely detectable. Accessibility imposes a second condition: the margin has to survive dichromatic simulation,

$$\min_{D \in \{prot,\, deut,\, trit\}} \Delta E_{00}(\, D(c_q), D(c_c)\,) \ge \Delta E^* \qquad (18)$$

with D the standard dichromat simulations [13]. Where Equation (18) cannot be met, the channel moves to a non-chromatic modality (weight, case, or a typographic mark) under the same $\chi$ with a different codomain, and the whole analysis carries over unchanged because no theorem below uses any property of colour other than the existence of the margin. This closes a limitation that the earlier version flagged and left open.

One honest caveat belongs here rather than in the discussion. Equations (17) and (18) make the channel more visible, and visibility to the candidate is visibility to a screenshot. Raising $\Delta E^*$ strengthens the guarantee against A1 and A2 and weakens nothing there, but it makes the separation trivially legible to A3 and A5. The channel is not a secret; it is a filter that certain capture routes destroy.

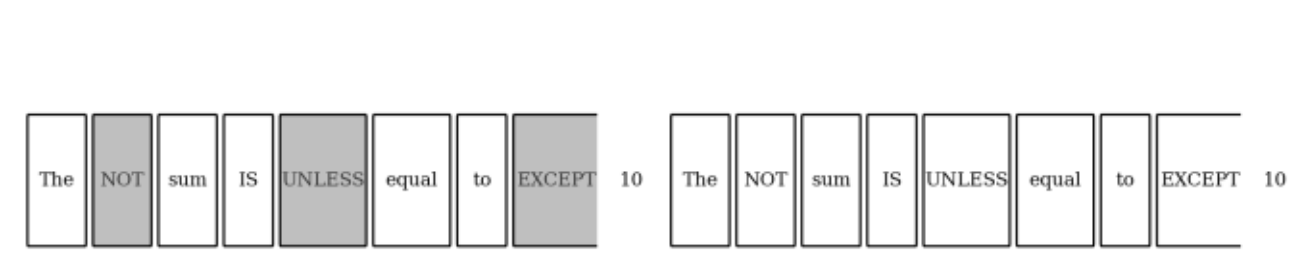

*Fig. 3. Dual-channel perception under separation: a legitimate candidate perceiving Γ filters the laminated stream preattentively, while a class A1 or A2 extraction channel sees only the undifferentiated token sequence.*

## VIII. Readability and Ambiguity Functionals

Two functionals quantify the trade-off that separated lamination is designed to exploit. The Human Readability Functional is

$$R_h = (1/n)\ \Sigma_{i=1..n}\ \alpha_i \tag{19}$$

where the perception indicator is

$\alpha_i = 1$ if authentic token i is correctly perceived;

$= 0$ otherwise (20)

The earlier statement that $R_h \to 1$ independently of density is true but empty, because it assumes exactly what it concludes. What makes it a claim about the world is the condition under which $\alpha_i = 1$ holds, and visual-search theory supplies it: separation by a single preattentive feature yields search time approximately flat in distractor count, whereas a below-threshold margin forces serial search and readability then falls with density [9], [10]. We therefore model perception as

$$\Pr[\ \alpha i = 1\ ] = 1 - \epsilon(\Delta E00),\ \text{with } \epsilon \text{ decreasing in } \Delta E00 \text{ and } \epsilon \to 0 \text{ for } \Delta E00 \geq \Delta E^* \tag{21}$$

and a companion search-cost term that Section IX prices,

$$Sc(\eta) = a + b\cdot\eta\cdot 1[\ \Delta E00 < \Delta E^*\ ] \tag{22}$$

so that above threshold the cost of camouflage to the candidate is a constant a and below threshold it grows linearly in η. Theorem 2 is stated against Equation (21) rather than against an assumption.

The Computational Ambiguity Functional needs a more substantial correction. It was previously defined as the excess unigram entropy $A_c = H(\Omega) - H(Q)$. That quantity is not monotone in camouflage density and can fall as tokens are added: inserting the same camouflage token repeatedly concentrates the token distribution and reduces $H(\Omega)$. It also measures the wrong thing, since the adversary's difficulty is uncertainty about Q, not the entropy of what is on screen. We redefine it as the conditional entropy of the authentic stream given the adversary view,

$$A_c = H(\ Q \mid V_A\ ) \tag{23}$$

For an A1 or A2 adversary, $V_A = \Omega$, and Theorem 3 gives the closed form

$$A_c = \log_2 C(n + m, m) \tag{24}$$

which is strictly increasing in m for fixed n, is exactly the quantity that Theorem 5 bounds, and reduces the two results to one. For an A3 adversary the view includes χ, and $A_c = 0$. For A4 and A5 the semantic prior reduces the effective count, which Theorem 6 bounds.

## IX. Fidelity, Obfuscation, and Plausibility

The original framework imposed a single Context Preservation Constraint, $Sim(Q, \Omega) \geq \theta$, on the grounds that the authentic content must remain recoverable and semantically intact. Recoverability is a separate matter from similarity, and once the two are separated the constraint turns out to point the wrong way. High $Sim(Q, \Omega)$ means the flat laminated field still conveys the item, which is exactly the property that lets a solver holding the extracted text answer it. The correct design has three constraints rather than one.

Fidelity. The legitimate rendering path must return the item exactly, not approximately:

$$\mathscr{R}(\ \Sigma\ ;\ \Gamma\ ) = Q \quad \text{(exact, not up to similarity)} \tag{25}$$

This is an identity guaranteed by Theorem 5, and it makes the similarity floor unnecessary for the purpose it was introduced to serve.

Obfuscation. The adversary view must not convey the item:

$$Sim(\ Q, \Omega\ ) \leq \theta_{obf} \tag{26}$$

with Sim instantiated by a sentence encoder such as Sentence-BERT [5] over contextual representations [4]. The inequality is a ceiling where [1] had a floor.

Plausibility. The laminated field must not be trivially separable by a language model, or Theorem 3 overstates the adversary's work by a wide margin. Let $p_{LM}$ be a reference model and define the per-position detectability

$$\delta_i = \mid \log p_{LM}(\Omega_i \mid \Omega_{<i}) - E_{w\sim Q}[\ \log p_{LM}(w \mid \Omega_{<i})\ ] \mid \tag{27}$$

The constraint is a ceiling on how far camouflage positions stand out,

$$(1/m)\ \Sigma_{i \in \Pi}\ \delta_i \leq \delta_{max} \tag{28}$$

Equations (26) and (28) are in tension by construction, which is the trilemma the designer actually faces: a camouflage stream distant enough to destroy $Sim(Q, \Omega)$ tends to be distant enough to be flagged by $p_{LM}$, and one bland enough to pass Equation (28) tends to leave the item readable. Section XV returns to this point, and we regard it as the main open problem the framework raises.

The intensity of camouflaging is controlled by the camouflage density

$$\eta = m / n \tag{29}$$

and the design problem is a constrained maximization rather than a search for a single crossing point,

$$\eta^* = \arg\max_{\eta \in F}\ [\ A_c(\eta) - \lambda\ S_c(\eta)\ ] \tag{30}$$

in which λ prices candidate effort and the feasible set is $F = \{\eta$ : Equations (26) and (28) hold, and $R_h(\eta) \geq 1 - \varepsilon\}$. Theorem 4 gives conditions for $\eta^*$ to exist and identifies when it sits on the boundary of F. Figure 4 shows the qualitative shape of the trade-off; the curves are schematic and carry no measured values.

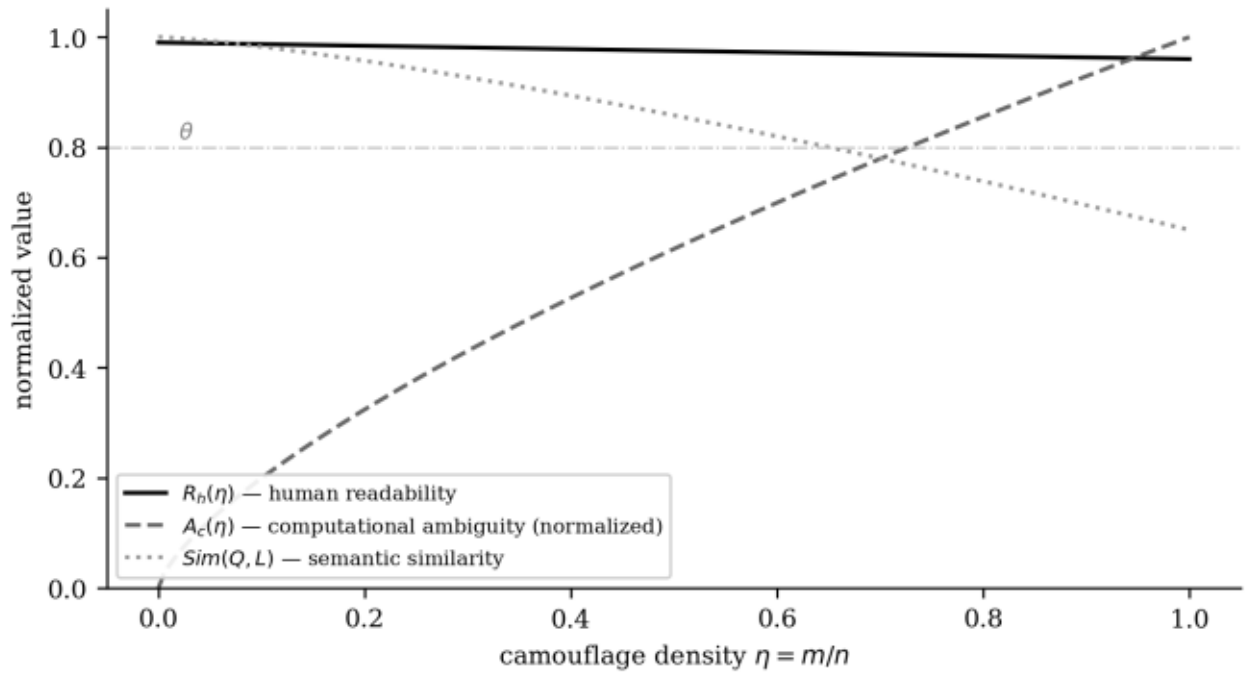

*Fig. 4. Schematic trade-off between human readability, computational ambiguity, and semantic similarity as functions of camouflage density η, under an above-threshold separation margin. Curves are illustrative; Section XIV specifies how they would be measured.*

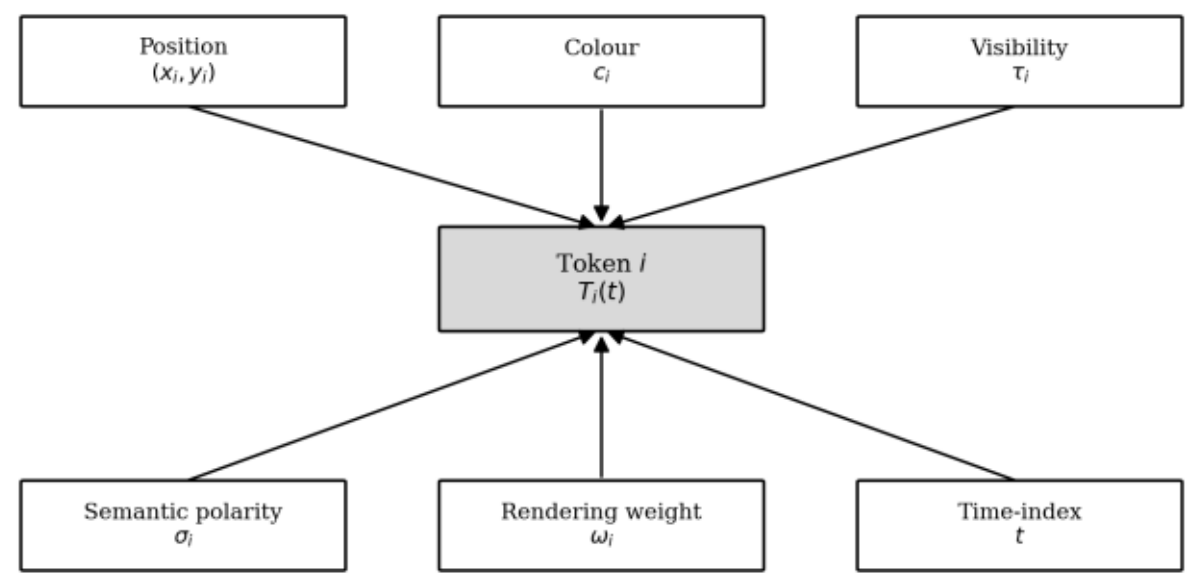

*Fig. 5. The Context Camouflage Tensor T_i(t): each rendered token carries position, separation value, visibility, semantic polarity, rendering weight, and a time index.*

## X. Dynamic Lamination and the Context Camouflage Tensor

Consistent with MARS's session-adaptive rendering philosophy, the camouflage stream need not be static. Dynamic lamination re-samples $\tilde{Q}$ over time,

$$\Omega(t) = \mathscr{L}( Q, \tilde{Q}(t); \Lambda(t) ) \qquad (31)$$

so that the camouflage layer evolves while the constraint set of Section IX continues to hold at every t. Every rendered token additionally carries a Context Camouflage Tensor recording its rendering state,

$$T_i(t) = \langle x_i, y_i, c_i, \tau_i, \sigma_i, \omega_i \rangle \qquad (32)$$

where $(x_i, y_i)$ is spatial position, $c_i$ the assigned separation value, $\tau_i$ visibility, $\sigma_i$ semantic polarity (authentic or inverted), and $\omega_i$ rendering weight.

Equation (31) is where the earlier framework had a genuine vulnerability, and we would rather state it than let a reader discover it. If Q is held fixed while $\Lambda(t)$ and $\tilde{Q}(t)$ are resampled every frame, then across k captures the authentic positions are identifiable as the ones whose token never changed. Randomizing the camouflage layer more aggressively makes the leak faster, not slower. Theorem 7 gives the decay rate. The condition under which temporal variation is safe is that the per-position temporal process be identically distributed for the two classes,

$$\text{law}( \Omega_i(t_{1:k}) \mid i \notin \Pi ) = \text{law}( \Omega_i(t_{1:k}) \mid i \in \Pi ) \qquad (33)$$

Equation (33) is not satisfied by resampling only the camouflage layer. Two designs do satisfy it in practice. Either hold both Π and $\tilde{Q}$ fixed for the lifetime of a session and rotate only across sessions, which reduces the leak to zero at the cost of intra-session adaptivity; or vary the authentic surface too, by resampling an equivalence-preserving surface form of each authentic token on the same schedule, which preserves adaptivity at the cost of a harder inversion operator. We recommend the first as the default and treat the second as future work.

## XI. Unified Rendering Equation

MCCT connects to the rendering dynamics of the original MARS framework through a unified rendering equation. Recall from [1] the per-word visibility function

$$\Psi_i(t) = \mathbb{1}[ \sin(2\pi f_i t + \varphi_i + \psi) - \tau \geq 0 ] \qquad (34)$$

and let $\delta_i(t)$ denote the positional-drift function governing token i's spatial rendering.

The previous version wrote the rendering as a scalar product $\Psi_i(t)\cdot\delta_i(t)\cdot\Omega_i\cdot C(w_i)$, which multiplies a binary indicator, a displacement, a token, and a colour. Those are not commensurable quantities and the expression has no defined value. The rendering is a map into a glyph state, so we write it as one:

$$\mathscr{R}_i(t) = \langle \Omega_i,\ p_i + \delta_i(t),\ \chi(i),\ \Psi_i(t)\cdot\omega_i \rangle \qquad (35)$$

whose four components are the glyph, the drifted position, the separation value, and the opacity. Only the last is a scalar, and the other three keep their own types. The rendered surface of Equation (2) is the collection $\{ \mathscr{R}_i(t) \}$, and the visible fraction at any instant is

$$\rho(t) = (1/n) \Sigma_{i \notin \Pi} \Psi_i(t) \qquad (36)$$

which is the parameter that Theorem 8 turns into a capture-count bound. Equation (35) is the point at which MCCT rejoins, and becomes a fully specified special case of, the general MARS Rendering Tensor formalism.

## XII. Theoretical Analysis

The following eight results characterize the joint behaviour of the constructs of Sections IV to XI. Theorems 1, 2, 4, and 5 revise results stated in the earlier version; Theorems 3, 6, 7, and 8 are new. Each is stated against a named adversary class from Table I.

***Theorem 1 (Separation-Conditioned Entropy Collapse).*** Let Σ be the rendered surface of Equation (2) and Γ the separation key. For an observer of Σ who resolves the separation channel, $H(Q \mid \Sigma, \Gamma) = 0$ exactly. For an adversary of class A1 or A2, whose view is Ω, $H(Q \mid \Omega) = A_c > 0$ whenever $m \geq 1$.

*Proof.* Resolving the channel yields, at every position i, the indicator $\mathbb{1}[\chi(i) = c_c]$ and therefore the set Π exactly. Deleting the positions in Π and reading the remainder in order returns Q, by the order-preserving piecewise definition of Equation (14). Q is thus a deterministic function of (Σ, Γ), and the conditional entropy of a deterministic function is zero. The adversary of class A1 or A2 receives Ω with χ discarded, so Π is not determined, and the residual uncertainty is $A_c$ by Definition (23), which Theorem 3 shows is positive for $m \geq 1$. ■

This strengthens the earlier statement, which carried a residual term ε for positional ambiguity within the authentic class. No such term is needed. Lamination is order-preserving, so once Π is known the authentic subsequence is recovered without ambiguity, and the collapse is exact rather than approximate.

***Theorem 2 (Readability and Ambiguity Decouple Above the Discriminability Threshold).*** If $\Delta E_{00}(c_q, c_c) \geq \Delta E^*$, then $E[R_h] = 1 - \epsilon$ with ϵ independent of η, and the search cost $S_c(\eta)$ of Equation (22) is constant in η, while $A_c(\eta)$ is strictly increasing in η. If $\Delta E_{00} < \Delta E^*$, decoupling fails: ϵ grows with m and $S_c$ grows linearly in η.

*Proof.* Under Equation (21) the indicators $\alpha_i$ are Bernoulli with a common failure probability $\epsilon(\Delta E_{00})$ that depends on the separation margin and not on the number of camouflage tokens, which is the defining property of a preattentive feature-search regime [9], [10]. Linearity of expectation applied to Equation (19) gives $E[R_h] = 1 - \epsilon$ for every η. With the indicator in Equation (22) false, $S_c(\eta) = a$. Strict monotonicity of $A_c$ in η follows from Theorem 3, since C(n+m, m) is strictly increasing in m for fixed $n \geq 1$. Below threshold the discrimination is no longer a single-feature pop-out, search becomes serial in the distractor count, ϵ acquires a dependence on m, and the second term of Equation (22) activates. ■

The distinction matters for deployment. The earlier version obtained decoupling by assuming $\alpha_i = 1$, which makes the conclusion true by construction. Here decoupling is a consequence of a physical condition on the rendered colours, and it is a condition that a platform can verify at render time.

***Theorem 3 (Closed-Form Computational Ambiguity).*** Let Π be drawn uniformly among the m-subsets of {1, …, n+m} and let the adversary view be Ω. Then $A_c = H(Q \mid \Omega) \leq \log_2 C(n + m, m)$, with equality when the candidate streams induced by distinct subsets are distinct. At unit density m = n, Stirling's approximation gives $A_c \approx 2n - \frac{1}{2}\log_2(\pi n)$.

*Proof.* Each candidate subset S of size m induces a candidate authentic stream $Q_S = (\Omega_i : i \notin S)$. By Equation (14) the true Q equals $Q_\Pi$. The posterior over candidates given Ω is the pushforward of the uniform prior on Π under $S \mapsto Q_S$. A uniform distribution on a set of size N has entropy $\log_2$ N, and pushforward under a map that is not injective can only merge outcomes and reduce entropy, so $H(Q \mid \Omega) \leq \log_2 C(n+m, m)$ with equality exactly when the map is injective. Repeated tokens in Ω are the only source of collisions. The unit-density expression follows from $C(2n, n) \approx 4^n / \sqrt{\pi n}$. ■

Two consequences are worth noting. Ambiguity grows linearly in item length at fixed density, roughly two bits per authentic token at η = 1, so longer items are intrinsically better protected. And repeated tokens are a leak: a camouflage stream that reuses the same few markers reduces $A_c$ below the bound, which is a second reason, alongside condition C3, to draw camouflage tokens from a wide distribution.

***Theorem 4 (Existence and Location of the Optimal Density).*** Let F be the feasible set of Equation (30). If F is non-empty then η* exists. If the separation margin is above threshold, so that $S_c$ is constant on F by Theorem 2, then η* = max F: the optimum lies on the boundary and at least one of Equations (26) and (28) is active at η*.

*Proof.* For fixed n the density η = m/n takes finitely many values in [0, 1], so F is a finite set and a maximum of any real objective over a non-empty finite set is attained. In the continuous relaxation used for tuning, $A_c(\eta)$ is continuous and strictly increasing and Sim(Q, Ω(η)) is continuous and non-increasing, so F is closed and bounded and therefore compact, and an upper semicontinuous objective attains its maximum on it. Above threshold the objective $A_c(\eta) - \lambda S_c(\eta)$ reduces to $A_c(\eta) - \lambda a$, which is strictly increasing, so its maximizer over F is sup F, which lies in F by closedness. Since $A_c$ increases without bound in m while Equations (26) and (28) are eventually violated, sup F is interior to [0,1] and is determined by whichever constraint fails first. ■

The earlier version treated η as ranging continuously over a compact interval and concluded existence from continuity alone. The finite-value argument is both simpler and correct, and the boundary result is the practically useful part: there is no interior optimum to search for above threshold, so tuning reduces to finding the largest density that still satisfies the obfuscation and plausibility ceilings. An interior optimum reappears only below threshold, where $\lambda S_c(\eta)$ grows with η and genuinely trades against $A_c$.

***Theorem 5 (Lamination Invertibility and Brute-Force Cost).*** Given (Π, Γ), $Q = \mathscr{L}^{-1}(\Omega; \Pi)$ exactly. Without them, an adversary of class A1 or A2 confronts a candidate set of size C(n+m, m), that is, 2 raised to the power $A_c$.

*Proof.* Invertibility is the construction in the proof of Theorem 1. The cardinality is immediate from Theorem 3. ■

The framing is deliberately modest and we want to be explicit about what it is not. This is an average-case entropy statement under a uniform prior over insertion sets. It is not a computational hardness result: there is no reduction to a problem believed to be hard, and none is claimed. Theorem 6 shows how far the bound falls against an adversary who does more than enumerate.

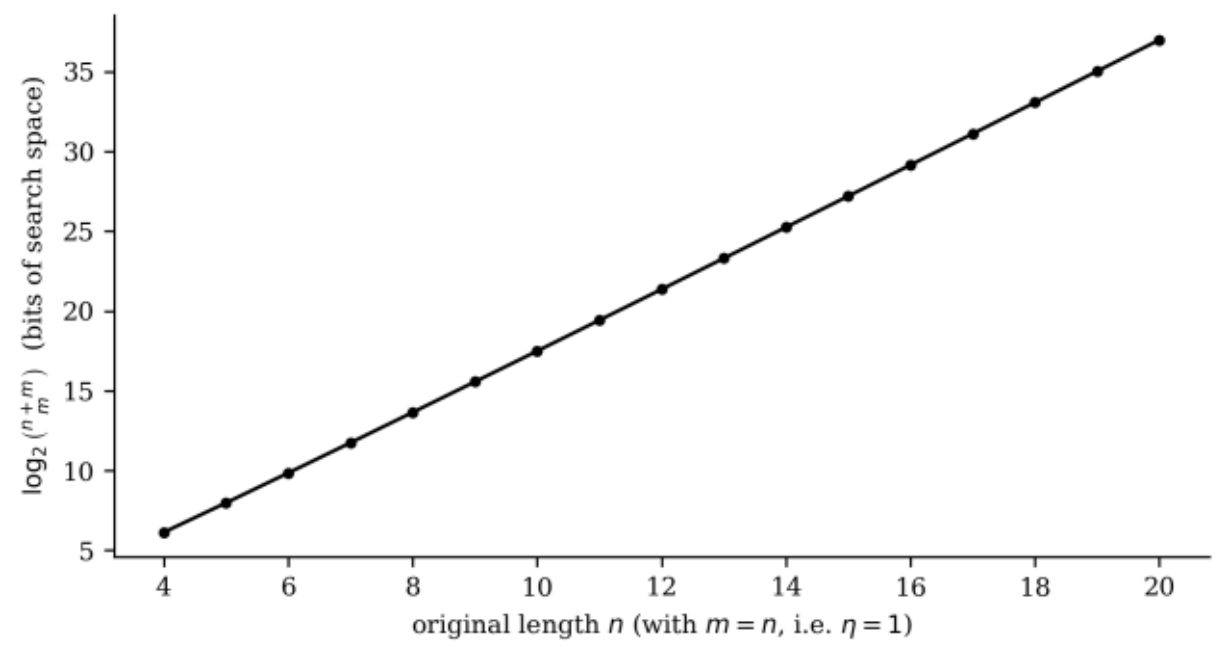

*Fig. 6. Growth of the candidate-set size C(n+m, m), in bits, as a function of authentic-stream length n at unit camouflage density (m = n). This is the quantity A_c of Theorem 3.*

***Theorem 6 (Degradation Under a Semantic Filter).*** Let an adversary of class A4 or A5 hold a reference model $p_{LM}$ and retain only candidates whose sequence probability exceeds a threshold, and let q be the acceptance rate of that test under the uniform prior over insertion sets. Then the residual ambiguity is $A_c^{LM} = A_c - \log_2(1/q)$. The plausibility constraint of Equation (28) is exactly a constraint on q.

*Proof.* The filter restricts the candidate set of Theorem 3 to its accepted subset, whose expected cardinality is $q \cdot C(n+m, m)$. Conditioned on acceptance, and absent further information distinguishing accepted candidates, the posterior is uniform on that subset, whose entropy is $\log_2(q \cdot C(n+m, m)) = A_c - \log_2(1/q)$. Camouflage tokens that are conspicuous under $p_{LM}$ drive q toward $1 / C(n+m, m)$ and hence $A_c^{LM}$ toward zero; camouflage tokens indistinguishable under $p_{LM}$ give $q \approx 1$ and leave $A_c$ intact. Since Equation (28) bounds the per-position detectability of camouflage tokens under the same model, it bounds q from below. ■

This is the result that connects the design constraints to the adversary that matters most in current practice. It also makes the trilemma of Section IX quantitative: Equation (26) pushes camouflage tokens away from the authentic distribution, which lowers q and therefore lowers $A_c^{LM}$, while Equation (28) pushes them back toward it, which raises q but raises Sim(Q, Ω) with it. There is no setting of the inversion operator that maximizes both, and we do not have a principled way to choose the exchange rate between them.

***Theorem 7 (Multi-Observation Leakage Under Frame-Granular Re-Lamination).*** Suppose Q is fixed within a session while camouflage tokens are redrawn independently at each of k rendered frames from a distribution P, and let $\gamma_k = \Sigma_w P(w)^k$, which equals 2 raised to the power $-(k-1)H_k(P)$ with $H_k$ the Rényi entropy of order k. An adversary who aligns the k captures and retains positions whose token is invariant across all of them retains every authentic position and, in expectation, $m_k = m \cdot \gamma_k$ camouflage positions. Residual ambiguity satisfies $A_c^{(k)} \leq \log_2 C(n + m_k, m_k)$, which decays geometrically in k.

*Proof.* An authentic position carries the same token at every frame and so survives the invariance test with probability one, giving n survivors. A camouflage position survives only if k independent draws from P coincide, which occurs with probability $\Sigma_w P(w)^k = \gamma_k$; linearity of expectation gives $m_k = m\, \gamma_k$ expected camouflage survivors. The surviving set contains Q as a subsequence together with at most $m_k$ impostor positions, and Theorem 3 applied to that reduced instance bounds the remaining uncertainty. Since $H_k$ is non-increasing in k and bounded below by the min-entropy $H_\infty(P)$, the decay rate lies between $H_\infty$ and $H_2$ bits per additional frame. ■

Figure 7 plots the decay for a 120-token item at unit density. At two bits of effective decay per frame, ambiguity falls from roughly 240 bits to under one bit within eight captures, and at four bits per frame within five. The direction of the effect is the point. Resampling the camouflage layer more aggressively raises the per-frame entropy $H_k$, which accelerates the leak rather than slowing it. Dynamic lamination as described in the earlier version therefore weakens the scheme against any adversary who captures more than once, and the invariance test costs the adversary nothing beyond alignment.

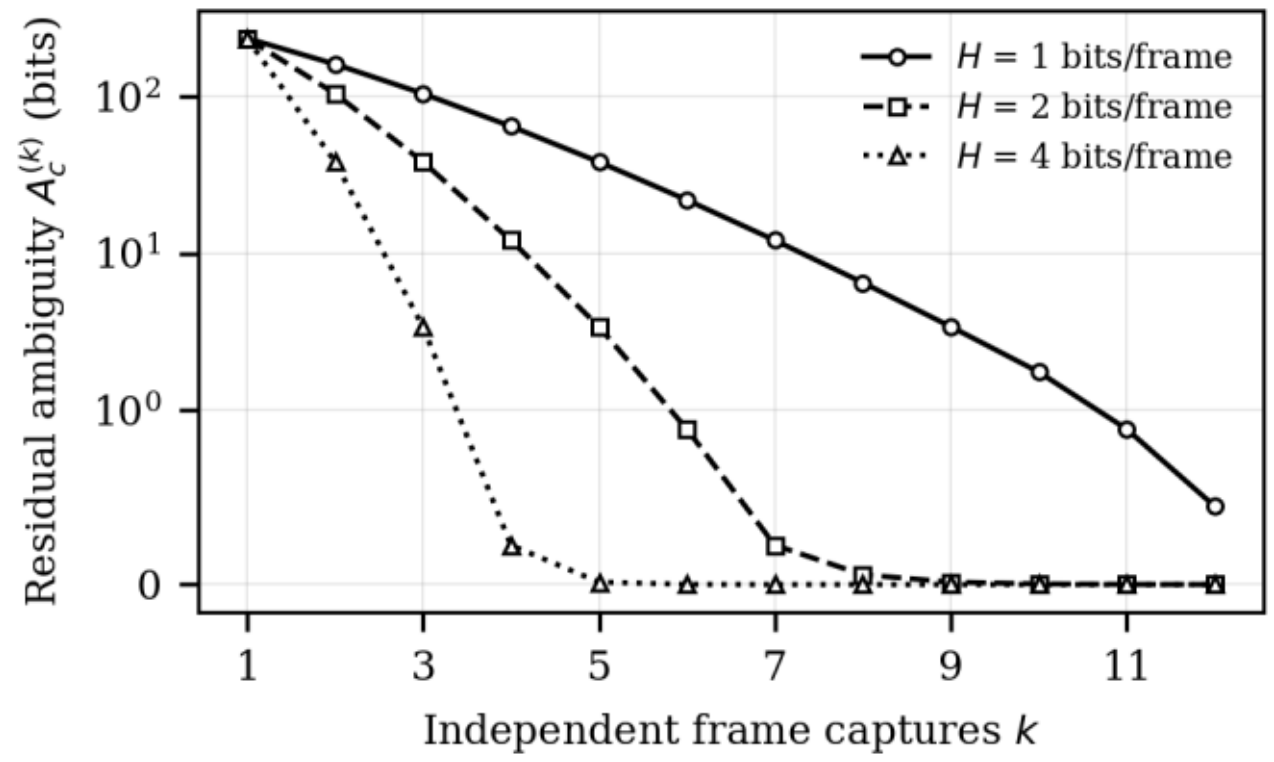

*Fig. 7. Residual ambiguity after k independent frame captures under frame-granular re-lamination, for a 120-token item at unit camouflage density, plotted from the bound of Theorem 7 at three effective decay rates. Curves are computed from the closed form and contain no measured data.*

The remedy is Equation (33). Holding both Π and $\tilde{Q}$ fixed for the lifetime of a session removes the invariance signal entirely, since nothing varies for the adversary to compare, and confines rotation to the session boundary where an attacker gains no repeated observations of the same item. We take this as the default configuration and treat frame-granular variation as unsafe unless the indistinguishability condition is verified.

***Theorem 8 (Capture Complexity Under Temporal Multiplexing).*** Let each frame reveal each authentic token independently with probability ρ, as in Equations (34) and (36). Then the expected number of captures required to observe every authentic token at least once is $E[k] = \Sigma_{j\geq 0} [\, 1 - (1 - (1-\rho)^j)^n \,] \approx \ln n / \ln(1/(1-\rho))$. This bound applies to every adversary class, including A5.

*Proof.* The first frame at which token i becomes visible is geometric with parameter ρ, and these are independent across tokens. The number of frames until all n have appeared is the maximum of n independent geometric variables, whose expectation is the stated sum by the tail-sum identity, with the classical extreme-value approximation $\ln n / \ln(1/(1-\rho))$ for large n. Any reconstruction of Q requires each of its tokens to have been observed at least once, so E[k] lower-bounds the expected captures needed for exact recovery. ■

Figure 8 shows the growth. The honest reading is that temporal multiplexing raises the cost of a screenshot attack from one capture to a modest number, and that the growth in item length is logarithmic rather than polynomial, so it is a friction rather than a barrier. Its value comes from composition with the behavioural domain B(t) of the MARS state: if each capture carries an independent detection probability $p_d$, the probability that a full reconstruction goes unnoticed falls as $(1 - p_d)^{E[k]}$, and it is that product, not the capture count alone, that constitutes the defence against A5.

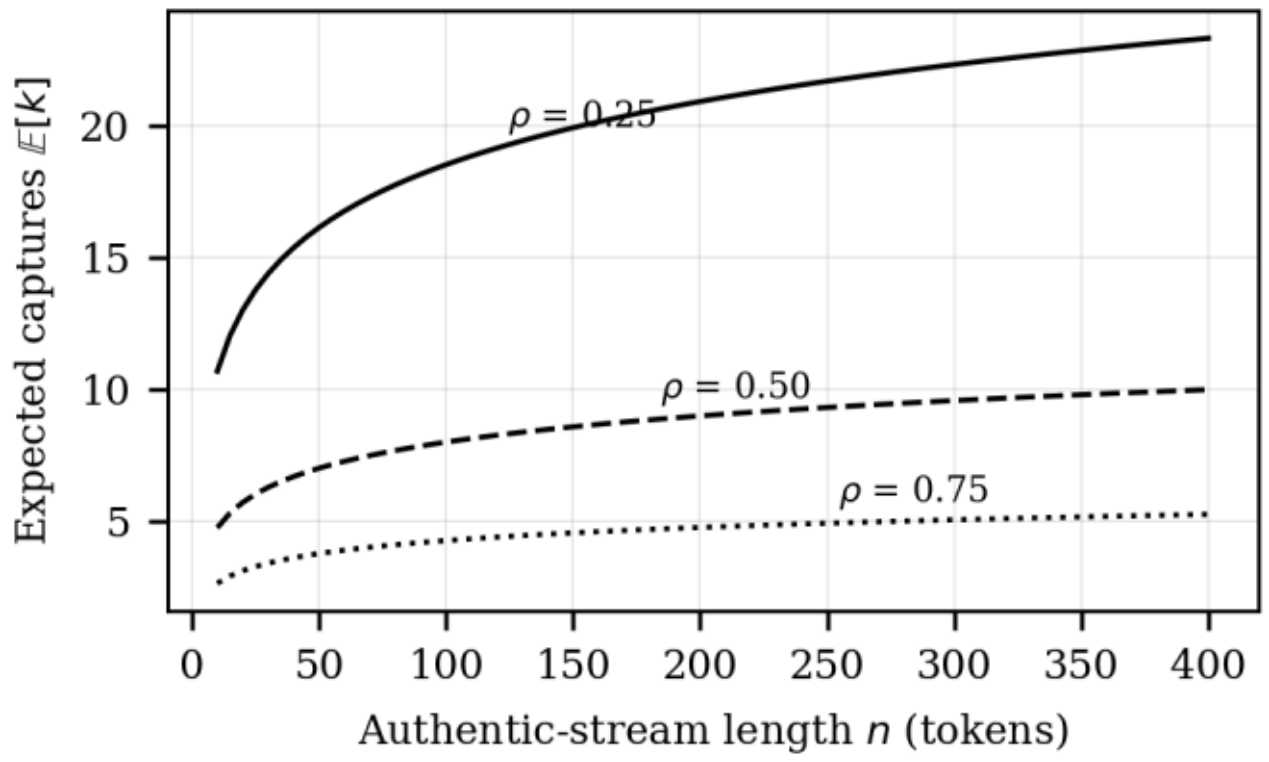

*Fig. 8. Expected number of captures required for complete observation of the authentic stream under temporal multiplexing, from the exact expression of Theorem 8, at three per-frame visible fractions ρ. Computed from the closed form; no measured data.*

## XIII. Algorithm and Complexity

The constructs of Sections V to VII compose into a single render-time procedure. Stating it makes the cost explicit and removes any impression that the framework requires expensive inference on the critical path.

```
Algorithm 1  Render a camouflaged item
Input : Q=(w_1..w_n), key K, item id, eta,
        margin dE*, ceilings th_obf, d_max
Output: rendered surface Sigma
1  m   <- round(eta*n)
2  Qt  <- Inversion(Q, m)      # C1-C3
3  Pi  <- PRF(K, id, 0)        # m-subset
4  Gam <- PRF(K, id, 1) s.t. (17),(18)
5  for i = 1..n+m:             # laminate
6      Om[i]  <- Qt[j(i)] if i in Pi
7                else Q[r(i)]
8      chi[i] <- c_c if i in Pi else c_q
9  if Sim(Q,Om) > th_obf
10    or Detect(Om,Pi) > d_max:
11     lower eta or resample Qt; goto 2
12 return Sigma <- {(Om[i],p_i,chi[i],Psi_i)}
```

Lamination is a single pass, so lines 5 to 8 cost $\Theta(n+m)$ time and $\Theta(n+m)$ space. Key derivation is $\Theta(m)$. The inversion operator dominates: with a lexical instantiation over WordNet it is $\Theta(m)$ lookups, and with a generative instantiation it is one forward pass over the item. The verification at lines 9 and 10 costs one sentence-encoder pass for Sim and one language-model pass for Detect. Since Q, $\tilde{Q}$, Π, and Γ all depend only on the item and the session key, the whole procedure runs once per item per session and the result is cached, so the per-frame render path carries only the $\Theta(n+m)$ evaluation of Equations (35) and (36). The retry loop at line 11 terminates because $A_c$ is monotone in η and η = 0 trivially satisfies both ceilings.

## XIV. Proposed Evaluation Protocol

This paper reports no experiments and we make no empirical claims. The theorems above are, however, stated so that each yields a falsifiable prediction, and we set out the protocol here so that a later study can be pre-registered against it rather than assembled after the fact.

Readability, testing Theorem 2. Candidates complete matched item sets at densities spanning $\eta \in [0, 1]$ under an above-threshold and a below-threshold margin. Primary outcomes are item-level accuracy and time to first keystroke. The prediction is that above threshold both are flat in η within noise, and that below threshold time grows linearly in η. Participants should include a colour-vision-deficient group under the non-chromatic channel of Section VII.

Extraction resistance, testing Theorems 3, 5, and 6. For each adversary class in Table I, the extracted view is passed to a solver and scored for item-level accuracy. The prediction is a large drop for A1 and A2, no drop for A3 absent styling countermeasures, and an intermediate drop for A4 and A5 that tracks the acceptance rate q of Theorem 6. Measuring q directly, by scoring candidate reconstructions under a held-out model, is the sharpest single test of the theory.

Leakage, testing Theorem 7. Under frame-granular re-lamination, apply the invariance test at k = 1, 2, 4, 8 captures and record the fraction of authentic tokens correctly identified. The prediction is that identification approaches unity within a number of captures set by the per-frame decay rate, and that holding Π and $\tilde{Q}$ fixed across the session drives it to chance.

Reporting. Effect sizes with confidence intervals, the full density grid rather than a selected operating point, and the measured $\Delta E_{00}$ of the rendered colours, which is the parameter on which the readability results depend and which is easy to leave unreported.

## XV. Discussion and Limitations

The constructs developed here reframe Context Camouflaging from a single black-box transformation Φ into an explicit pipeline whose readability and ambiguity properties are separately tuneable and, above the discriminability threshold, decoupled. The practical implication is that an institution can raise computational ambiguity against class A1 and A2 extraction by raising η without a corresponding cost to legitimate candidates, provided the separation margin of Equations (17) and (18) is met at render time.

The scope of that claim is narrower than the earlier version implied, and the narrowing is the main contribution as much as the new theorems are. Colour separation defends the text-extraction channel. It does not defend against a scripted reader of the document object model, which sees the styling, and it does not defend against a vision-language model given a screenshot, which sees both the colour and enough semantics to make Theorem 6 bite hard. Against those classes the defence has to come from the temporal domain, and Theorem 8 prices what that buys: a logarithmic increase in capture count, valuable mainly because each capture carries detection risk in the behavioural domain. A reader who takes away only one thing should take away that the guarantee is channel-specific.

Several limitations follow. First, the trilemma of Section IX has no principled resolution in this paper. Equations (26) and (28) pull in opposite directions, Theorem 6 shows the exchange rate matters, and we have no method for setting it beyond tuning against a held-out model, which invites overfitting to that model.

Second, the Context Inversion Operator is treated abstractly. Conditions C1 to C3 constrain it but do not construct it, and whether a lexical instantiation over WordNet [3] or a learned generative model can satisfy all three at a useful density is an empirical question this paper does not settle.

Third, Theorem 3 assumes a uniform prior over insertion sets. A biased sampler, a predictable position map, or heavy reuse of camouflage tokens each reduce $A_c$ below the closed form, and the reduction is silent: nothing in the rendered output reveals it. Auditing the sampler is therefore a deployment requirement rather than an implementation detail.

Fourth, accessibility. Equation (18) states the condition but a platform still has to meet it under arbitrary user stylesheets, high-contrast modes, and screen readers, the last of which will read the laminated field in full unless the camouflage positions are marked for exclusion, which in turn hands class A3 exactly the signal it needs. We have no clean answer to that conflict and regard it as the sharpest practical objection to the approach.

## XVI. Conclusion

This paper extended the Context Camouflaging Operator of the MARS assessment-resilience framework into the Multi-Layer Context Camouflaging Theory, a specified mathematical treatment of semantic superposition for malpractice-resilient online assessment. Six coupled constructs replace the single coarse transformation of the original formulation, and eight theorems characterize their joint behaviour under an explicit adversary model.

Three of the revisions change what the framework claims rather than only how it is stated. The preservation constraint became a fidelity identity together with obfuscation and plausibility ceilings, because a similarity floor asks the laminated field to keep conveying the item it is meant to hide. Computational ambiguity became a conditional entropy with the closed form $\log_2 C(n+m, m)$, which is monotone in density where the previous unigram measure was not. And frame-granular dynamic lamination turned out to leak geometrically against a multi-capture adversary, so the safe default is session-granular rotation.

Equation (35) reconnects the theory to the MARS Rendering Tensor formalism, with the type error of the earlier scalar product removed, so Context Camouflaging is available as a specified layer within the larger system model. The open problems we would most like to see addressed are the exchange rate between obfuscation and plausibility in Theorem 6, a construction for the inversion operator satisfying conditions C1 to C3 at useful density, and the conflict between screen-reader accessibility and resistance to class A3.

***Table II***

**PRINCIPAL SYMBOLS USED IN MCCT**

| Symbol | Description | Unit / Domain |
|---|---|---|
| $Q$, $V$ | Authentic semantic stream / vocabulary | Token seq. / set |
| $\tilde{Q}$ | Context-camouflage stream | Token sequence |
| $\mathcal{J}$ | Context Inversion Operator | Mapping |
| $\Omega$ | Laminated (superposed) field | Token sequence |
| $\Sigma(t)$ | Rendered surface (new) | Attributed sequence |
| $X_A$, $V_A$ | Extraction channel / adversary view (new) | Projection / seq. |
| $\mathcal{L}(\cdot;\Lambda)$ | Contextual Lamination Operator | Mapping |
| $\Lambda = (\pi,\kappa,\chi)$ | Lamination control law | Function |
| $\Pi$ | Insertion set (camouflage positions) | Index set |
| $\chi(i)$, $\Gamma$ | Separation function / session key | Mapping / key |
| $c_q$, $c_c$ | Authentic / camouflage channel value | Colour value |
| $\Delta E_{00}$, $\Delta E^*$ | CIEDE2000 difference / margin (new) | CIELAB units |
| $R_h$, $\alpha_i$ | Human Readability Functional / indicator | 0–1 / binary |
| $S_c(\eta)$ | Visual search cost (new) | Time |
| $A_c$ | Computational Ambiguity, $H(Q \mid V_A)$ (revised) | Bits |
| $\mathrm{Sim}(\cdot)$, $\theta_{obf}$ | Semantic similarity / obfuscation ceiling | 0–1 |
| $\delta_i$, $\delta_{max}$ | Detectability under $p_{LM}$ / ceiling (new) | Log-prob |
| $q$ | Semantic-filter acceptance rate (new) | 0–1 |
| $\eta$, $m$, $n$ | Camouflage density / camouflage and authentic counts | Ratio / counts |
| $T_i(t)$ | Context Camouflage Tensor of token i | Tensor |
| $\Psi_i(t)$, $\delta_i(t)$ | Temporal visibility / spatial drift (from [1]) | Binary / scalar |
| $\rho(t)$ | Visible fraction per frame (new) | 0–1 |
| $\mathcal{R}_i(t)$ | Per-token rendering state (revised) | Tuple |

***AI Use Disclosure:*** *Portions of this manuscript were prepared with the assistance of generative artificial-intelligence tools for language refinement, formatting, and figure drafting. All scientific concepts, mathematical formulations, algorithms, theoretical propositions, analyses, and conclusions were conceived, verified, and approved by the authors, who accept full responsibility for the content of this manuscript [14].*